\documentclass[a4paper,fleqn]{cas-dc}
\renewcommand{\printorcid}{}

\usepackage[numbers]{natbib}

\def\tsc#1{\csdef{#1}{\textsc{\lowercase{#1}}\xspace}}
\tsc{WGM}
\tsc{QE}

\begin{document}
\let\WriteBookmarks\relax
\def\floatpagepagefraction{1}
\def\textpagefraction{.001}

\shorttitle{Unsupervised Person Re-Identification: A Systematic Survey of Challenges and Solutions}    

\shortauthors{X. Lin, P. Ren et al.}  

\title [mode = title]{Unsupervised Person Re-Identification: A Systematic Survey of Challenges and Solutions}  



%

\author[1]{Xiangtan Lin}



\ead{xiangtan.lin@gmail.com}


\credit{Conceptualization, Investigation, Formal analysis, Writing - original draft}

\affiliation[1]{organization={Monash University},
            addressline={Wellington Road}, 
            city={Clayton},
            postcode={3800}, 
            state={Victoria},
            country={Australia}}

\author[2]{Pengzhen Ren}


\ead{pzhren@foxmail.com}


\credit{Formal analysis, Writing - review \& editing}

\affiliation[2]{organization={Northwest University},
            city={Xi'an},
            state={Shaanxi},
            country={China}}

\author[1]{Chung-hsing Yeh}


\ead{Chung-hsing.Yeh@monash.edu}


\credit{Supervision, Writing - review \& editing}


\author[3]{Lina Yao}


\ead{lina.yao@unsw.edu.au}


\credit{Writing - review \& editing}

\affiliation[3]{organization={University of New South Wales},
            city={Sydney},
            postcode={2052}, 
            state={New South Wales},
            country={Australia}}

\author[4]{Andy Song}


\ead{andy.song@rmit.edu.au}


\credit{Writing - review \& editing}

\affiliation[4]{organization={RMIT University},
            addressline={124 La Trobe St}, 
            city={Melbourne},
            postcode={3000}, 
            state={Victoria},
            country={Australia}}

\author[4]{Xiaojun Chang}

\cormark[1]


\ead{xiaojun.chang@rmit.edu.au}


\credit{Conceptualization, Supervision, Funding acquisition, Resources, Writing - review \& editing}


\cortext[6]{* Corresponding author}



\begin{abstract}
Person re-identification (Re-ID) has been a significant research topic in the past decade due to its real-world applications and research significance. While supervised person Re-ID methods achieve superior performance over unsupervised counterparts, they can not scale to large unlabelled datasets and new domains due to the prohibitive labelling cost. Therefore, unsupervised person Re-ID has drawn increasing attention for its potential to address the scalability issue in person Re-ID. Unsupervised person Re-ID is challenging primarily due to lacking identity labels to supervise person feature representation learning. The corresponding solutions are diverse and complex, with various merits and limitations. Therefore, comprehensive surveys on this topic are essential to summarise challenges and solutions to foster future research. Existing person Re-ID surveys have focused on supervised methods from classifications and applications. Still, they lack detailed discussion on how the person Re-ID solutions address the underlying person Re-Id challenges. This survey review recent works on unsupervised person Re-ID from the perspective of challenges and solutions. Specifically, we provide an in-depth analysis of highly influential methods considering the four significant challenges in unsupervised person Re-ID: 1) lacking ground-truth identity labels to supervise person feature learning; 2) learning discriminative person features with pseudo-supervision; 3) learning cross-camera invariant person features and 4) the domain gap between datasets. We summarise and analyze evaluation results and provide insights on the effectiveness of the solutions. Finally, we discuss open issues and suggest some promising future research directions.
\end{abstract}



\begin{keywords}
Person re-identification
\sep  
Unsupervised person re-identification
\sep
Deep learning
\sep
Feature representation learning
\end{keywords}

\maketitle

\section{Introduction}

Person re-identification (Re-ID) aims to find the query person from a collection of person images or videos captured by non-overlapping cameras in a distributed multi-camera system~\cite{LiuCCPZYH20,LiLCYPZ19,ChengGCSHZ18}. Person Re-ID has wide real-life applications, such as criminals search, multi-camera tracking, missing person search etc. The essential task of person Re-ID is to learn the discriminative person features and associate the query with the best matching person in the gallery images or videos.  Historically, early person Re-ID models exploited handcrafted person features such as colours and textures and focused on metric learning to align the query person with the most similar person in the gallery images~\cite{LiuCCY18,ChengCLHGZ17,LiuC0Y17}. However, handcrafted features are ineffective in large-scale applications because they can not capture the dynamic person features in an extensive collection of gallery images. Therefore, the handcrafted feature engineering approach was superseded by CNN-based deep learning since 2014 ~\cite{li_deepreid_2014}  as the latter dominated computer vision research. 

Most deep learning person Re-ID models are trained with labelled data in a single domain. With CNN-based deep learning, supervised person Re-ID has achieved impressive accuracy in the commonly used benchmark datasets. In supervised person Re-ID, with sufficient pairwise labelled data, researchers have focused on designing novel network structures and efficient loss functions to learn cross-camera discriminative feature representation, which greatly improves person Re-ID performance. For instance, the Rank-1 benchmark of the single query search on the Market-1501~\cite{zheng_scalable_2015} dataset has increased from 44.4\% ~\cite{zheng_scalable_2015} at the time of release to 98.5\% ~\cite{liu_unity_2020} in 2020. The Rank-1 evaluation result of the DukeMTMC-Re-ID~\cite{zheng_unlabeled_2017} dataset has been lifted from only 30.8\% ~\cite{zheng_unlabeled_2017} in 2017 to over 95\%~\cite{liu_unity_2020} in 2020. 

The deep learning-based methods are data-driven and require many labelled data, which incur significant labour and computational costs for annotating data. Therefore supervised person Re-ID can not scale to large-scale datasets with prohibitive labelling costs. Moreover, those costly models trained on one dataset can not be directly deployed to other environments on the unlabeled datasets due to significant photometric and geometric variations between datasets, hindering the person Re-ID performance. To address the scalability issue of supervised person Re-ID, in recent years, researchers have been focusing on unsupervised person Re-ID to train Re-ID models with abundant unlabelled data. Unsupervised person Re-Id is not a new research topic in the computer vision community. Early attempts using handcrafted features and silence learning ~\cite{wang_unsupervised_2014} failed to produce satisfactory Re-ID results due to limited discriminative power in the person feature descriptors. It is the deep learning-based approach that delivers steady improvement in unsupervised person Re-ID performance. Therefore, this survey focuses on deep learning-based methods in unsupervised person Re-Id.

\subsection{Motivation and Related Surveys}
The performance of unsupervised person Re-ID is generally inferior to supervised person Re-ID due to lacking pairwise labelled data to learn camera-invariant person feature representation. In recent years, unsupervised person Re-ID has remarkably closed the performance gap with supervised person Re-ID. Therefore, unsupervised person Re-ID has drawn increasing attention in the last couple of years due to its scalable applications. A d diversity of solutions are proposed with various merits and limitations. To foster future research in this area, systematic surveys of unsupervised person Re-ID methods are essential. Recent person Re-ID surveys have paid little attention to unsupervised person Re-ID methods and lack comprehensive and in-depth analysis of the challenges and solutions. Only a few image-based unsupervised person Re-ID works are covered in recent person Re-ID surveys. To fill the gap, we conduct a comprehensive review of the diverse solutions of unsupervised person Re-ID covering both image-based and video-based methods. Specifically, We surveyed recently published and pre-print publications of unsupervised person Re-Id from top conferences and journals. Most current Re-ID surveys review methods from the perspective of applications and solutions that lack in-depth analysis of the rationals behind the solutions. To inspire new ideas to further advance unsupervised person Re-ID, we discuss the common challenges of unsupervised person Re-ID and provide insights on how the challenges are addressed in various solutions with convincing evaluation results. We summarise the main differences between this survey and related deep learning-based person Re-ID surveys in Table \ref{table:surveydiff}.

\subsection{Contribution}

Most existing Re-ID surveys emphasize the classification of diverse person Re-ID methods but lack cohesive analysis of how the solutions correspond to the underlying challenges. This survey aims to analyze the unsupervised person Re-ID methods from the perspective of challenges and solutions. Therefore, the rationals behind the solutions can be easily grasped to inspire new ideas. Specifically, We surveyed recently published and pre-print publications of unsupervised person Re-ID from top conferences and journal articles. We discuss the methods from the challenges and solutions perspective. We summarise and analyze evaluation results accordingly. We discuss open issues that lead to promising future research directions with an in-depth judgment of the challenges and corresponding solutions. The main contributions of this survey are summarised as follows:
\begin{itemize}
\item We emphasize unsupervised person Re-ID, which addresses the scalability issue of person Re-ID. This is the first survey dedicated to unsupervised person Re-ID, as far as we know.  
\item We analyze unsupervised person Re-ID methods from the perspective of challenges and solutions. Most current Re-ID surveys review methods from the perspective of applications and solutions that lack in-depth analysis of the rationals behind the ideas. To foster new ideas, we discuss the main challenges of unsupervised person Re-ID and provide insights on how the challenges are addressed.
\item We summarise and analyze existing unsupervised person Re-ID methods' performance and provide insights on promising future research directions.
\end{itemize}

\begin{table} [t]
\caption{Summary of the main differences between the related person Re-ID surveys and this survey.}
\renewcommand{\arraystretch}{1}
\begin{center}
\resizebox{\columnwidth}{!}{
\begin{tabular}{l|l|l}
\toprule
{\bf \huge Survey} & {\huge Covering} & {\huge Analysis} \\

\midrule
\parbox[c]{0.8in}{\huge~\cite{zheng_person_2016}} & \parbox[c]{1.5in}{\huge Handcrafted person Re-ID

Deep learning person Re-ID} & \parbox[c]{4.5in}{\huge Applications and methods} \\
\midrule
\parbox[c]{0.8in}{\huge ~\cite{chahar_study_2017,wang_survey_2018,lavi_survey_2018,wu_deep_2019,almasawa_survey_2019,ye_deep_2021,wang_beyond_2020,leng_survey_2020}} & \parbox[c]{1.5in}{\huge Deep learning person Re-ID

(mainly supervised)
} & \parbox[c]{4.5in}{\huge Applications and methods} \\
\midrule

\parbox[c]{0.8in}{\huge Ours} & \parbox[c]{1.5in}{\huge Unsupervised person Re-ID

Image-based and video-based

} & \parbox{4.5in}{\huge (Challenges: {\bf Solutions})

Lacking ground-truth identity labels:

{\bf Pseudo-label estimation}

Learning discriminative person features: 

{\bf Deep feature representation learning}

Learning camera-invariant person features

{\bf Camera-aware invariance learning}

Domain gap between datasets: 

{\bf Unsupervised domain adaptation}
} \\
\bottomrule
\end{tabular}
}
\label{table:surveydiff}
\end{center}
\end{table}

\subsection{Organisation of the Survey}

In this survey, we first overview the person Re-ID problem in Section \ref{overview} to set the context and highlight the essence of unsupervised person Re-ID. We then conduct a comprehensive analysis of unsupervised person Re-ID from the perspective of challenges and solutions in Section \ref{main}. After that, in Section \ref{evaluation}, we analyze the performance of representative solutions in response to the challenges. In Section 5, we discuss open issues that lead to promising future directions of unsupervised person Re-ID. Finally, we conclude the survey with the key takeaway messages in Section \ref{conclusion}.

\section{Overview of Person Re-Identification}\label{overview}

Person re-identification has been a significant research topic for over a decade due to its real-life applications, such as criminals searches, missing person searches etc. Person Re-ID heavily relies on visually similar person appearance from disjoint camera views to match the query person with the person images. Photometric and geometric variations in person appearance across cameras pose the greatest challenge. Most person Re-ID works have concentrated on supervised learning in various settings on small to large datasets. Occluded person re-identification (Re-ID) aims to match partial and whole person images across camera views~\cite{wang_high-order_2020,gao_pose-guided_2020,miao_pose-guided_2019,huang_adversarially_2018,vedaldi_guided_2020}. Long-term person Re-Id~\cite{qian_long-term_2020} aims to find the query person over a long period with changing clothes~\cite{qian_long-term_2020}. Homogeneous person Re-ID for single modality data such as Image-based and video-based person Re-ID have been wildly studied due to their simplicity. Heterogeneous person Re-ID aims to find the query person in cross-modality data~\cite{choi_hi-cmd_2020,lu_cross-modality_2020}, such as RGB-infrared person Re-ID~\cite{wu_rgb-infrared_2017}.  

\begin{figure*}[t]
\begin{center}
\includegraphics [width=.75\linewidth]{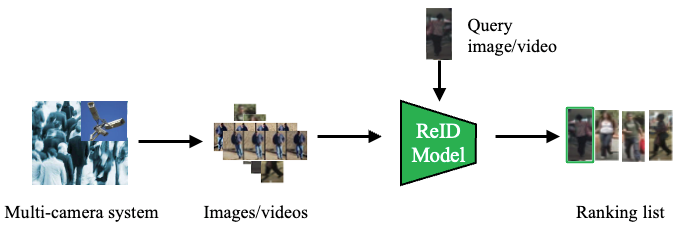}
\end{center}
  \caption{ A typical person re-identification system. A person Re-ID dataset is typically collected from a multi-camera system and curated for training and testing a Re-ID model. During training, the Re-ID model is to learn person feature representation. During testing, a query person is given to find the matching person in the gallery images. The output is a ranked list of the best matching person.}
\label{fig:overview}
\end{figure*}

Person Re-ID aims to find the query person from a collection of person images or videos captured by multiple disjoint cameras. Person Re-ID is an application of a multi-camera system from where the training and testing images/videos are sourced. The core part of the Re-ID system is to learn a Re-ID model to extract discriminative person features, which minimizes the distance with the query person. A typical person Re-ID system is illustrated in Figure \ref{fig:overview}. The overall objective of the Re-ID model can be expressed as:
\begin{equation}
\label{eq:reidobjective}
\min_{\theta,w} \sum_{i=1}^N \ell(f(w;\phi(\theta;x_i)),y_i).
\end{equation}
where $x_i$ is the person feature vector, $\phi(.)$ is a function to learn the feature representation of $x_i$, $\theta$ are the learnable parameters in $\phi(.)$. $f(.)$ is the classification function that takes input the person feature representation $\phi(.)$ and outputs the classification label. $w$ are trainable weights in $f(.)$. $y_i$ is the ground-truth identity label to supervise learning the feature extraction function $\phi(.)$. $\ell$ is the loss between the project label $f(.)$ and the ground-truth $y_i$. Therefore, the overall objective is to minimise the classification losses between predicted and ground-truth identity labels.

As illustrated in Equation \ref{eq:reidobjective}, the core part of the person Re-ID model system is to learn person feature representations so that two people have similar features can be regarded as the same identity. To learn suitable person feature representations, historically, prior to CNN-based deep learning, numerous hand-crafted features are exploited for person Re-ID, in particular the color and texture features, such as HSV color histogram ~\cite{farenzena_person_2010,li_locally_2013}, LAB color histogram~\cite{zhao_unsupervised_2013}, LBP histogram ~\cite{li_locally_2013}, Gabor features ~\cite{li_locally_2013} and SIFT ~\cite{zhao_unsupervised_2013}. In a handcrafted Re-ID system, such as LOMO ~\cite{liao_person_2015} and BOW ~\cite{zheng_scalable_2015}, it's critical to learn a suitable distance metric to close the gap between the query person and the handcrafted features since the handcrafted features can not capture the dynamic feature variance in the sample space. Therefore, the handcrafted methods are only suitable for small datasets and fail to fully exploit the data distribution to learn the appropriate feature representations for large-scale datasets.

Deep learning-based methods ~\cite{li_harmonious_2018,si_dual_2018,zhao_deeply-learned_2017,zhou_point_2017,li_person_2017} for feature extraction have shown substantial advantage over hard-crafted features because they fully exploit the data distribution of dynamic person features. Since 2014, deep learning-based supervised Re-ID methods have matured and have achieved over 95\% Rank-1 accuracy in several Re-ID datasets. Researchers have been working on novel network structures and effective training algorithms, coupled with effective global and local person features ~\cite{li_learning_2017,zhao_spindle_2017} and efficient loss functions etc. For example, a Domain Guided Dropout layer ~\cite{xiao_learning_2016} is proposed to illuminate useless neurons to learn global person representations from multiple domains. A multi-scale triplet CNN ~\cite{liu_multi-scale_2016} is proposed to extract multi-scale person features with a comparative similarity loss on triplet sample. ~\cite{mclaughlin_recurrent_2016} proposed an RNN architecture to exploit spatial and temporal information in videos to learn person features for video-based person Re-ID. ~\cite{li_person_2017} jointly learns local and global features with multiple classification losses.

\begin{figure*}[t]
\begin{center}
\includegraphics[width=\textwidth]{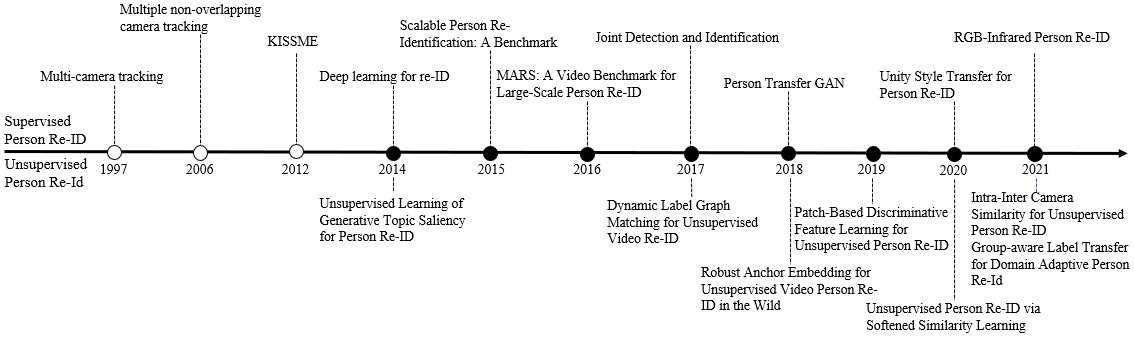}
\end{center}
\caption{Timeline of person Re-ID with representative papers. Above the timeline are some representative supervised person Re-ID works. Below the line are unsupervised person Re-ID works. Shallow dots indicate hand-crafted person Re-ID. Solid dots are deep learning-based person Re-Id.}
\label{fig:timeline}
\end{figure*}

Although these methods perform well on the single domain dataset when directly testing a model trained with the source dataset on the new target dataset, performance would drop dramatically, simply because the model hasn't seen the variations of the person features in the new domain. For a Re-ID model to generalize to a new environment, some works such as DIMN~\cite{song_generalizable_2019} is trained with several publicly labelled datasets. Person representation is averaged over multiple representations hoping that it would work in the unseen domain. DIMN deliberately uses large datasets as the source domains and the smaller ones as target domains. However, this doesn't address the labelled data scarcity problem. Testing on small datasets may work well as the model has seen similar features in the large dataset. Still, it doesn't guarantee that the model would generalize to other large datasets with significant variance in feature space. 

This generalization performance gap has led many researchers to unsupervised person Re-ID. Unsupervised person Re-ID has been investigated around the same time as supervised person Re-ID. Early works before the deep learning era mainly focus on how to construct effective feature representations manually. Traditional unsupervised person re-id studies have mainly focused on feature
engineering~\cite{farenzena_person_2010,gray_viewpoint_2008,zhao_unsupervised_2013}, which design appropriate handcrafted features using prior expert knowledge. Deep learning-based unsupervised person Re-ID has advanced significantly over the past few years, and several notable works have greatly advanced unsupervised person Re-ID performance. For instance, GLT~\cite{zheng_group-aware_2021} combines the pseudo-label prediction and Re-ID representation learning in one unified optimization objective. The holistic and immediate interaction between these two steps in the training process can significantly help the unsupervised person Re-ID task. The historical evolvement of person Re-ID is presented in Figure \ref{fig:timeline} with several representative works.
\begin{figure*}[t]
\begin{center}
\includegraphics [width=.75\linewidth]{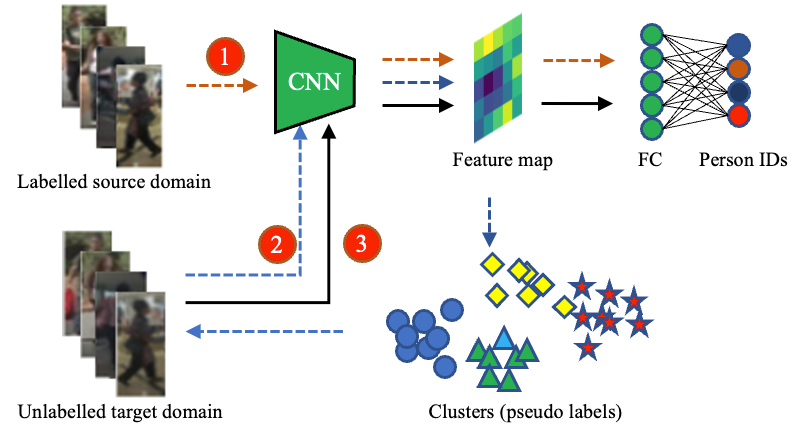}
\end{center}
  \caption{Illustration of unsupervised cross-domain person Re-ID. 1. Pre-train model with labelled source data. 2. Use the pre-trained model to extract and cluster features from unlabelled target data to obtain pseudo-labels. 3. Adapt the pre-trained model to target domain with pseudo-labels.}
\label{fig:clustering}
\end{figure*}
\section{Unsupervised Person Re-Identification}\label{main}

In this section, we review unsupervised person Re-ID methods from the challenges and solutions perspective. We first discuss the application scenario of unsupervised person Re-ID in unsupervised cross-domain person Re-ID and fully unsupervised person Re-ID based on whether a labelled source dataset is utilized. We then summarise the main challenges in unsupervised person Re-ID and analyze the corresponding solutions. Unsupervised person Re-ID has been studied in both image-based and video-based person Re-ID. Image-based unsupervised person Re-ID has been the mainstream research which is an extension of the image retrieval problem. Video-based unsupervised Re-ID has been gaining popularity in the past few years due to the rich temporal information in the video, which provides additional guidance in camera-invariant feature learning.

\begin{figure*}[t]
\begin{center}
\includegraphics [width=.75\linewidth]{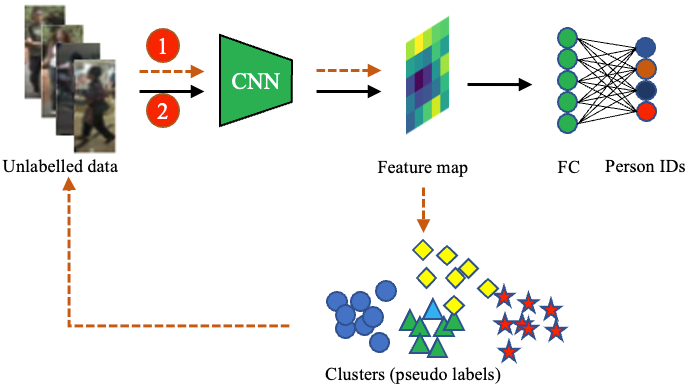}
\end{center}
  \caption{Illustration of fully-unsupervised person Re-ID. 1. Extract and cluster features from unlabelled target data to obtain pseudo-labels. 2. Train model with unlabeled data and pseudo-labels.}
\label{fig:fullyclustering}
\end{figure*}

CNN-based deep learning person Re-ID works ~\cite{sun_beyond_2018,wang_learning_2018} have achieved impressive Re-ID accuracy, but their success is largely dependent on sufficient annotated data that incur high labelling costs. In contrast, unlabeled person images are easy to collect from video systems, fostering the study of unsupervised Re-ID. In the current literature, unsupervised person Re-ID has two settings depending on whether using extra labelled data. It is generally perceived that it's difficult to learn camera-invariant person features without pairwise labels for the person identities appear in multiple camera views. Therefore, most works pre-train a Re-ID model on the labelled source dataset and adapt to the unlabelled target dataset, namely unsupervised cross-domain person re-identification ~\cite{wei_person_2018,zhong_invariance_2019,ge_mutual_2020} as illustrated in Figure \ref{fig:clustering}. Another group of unsupervised person Re-ID works directly with the unlabelled dataset without utilizing any labelled data, which is named fully unsupervised person Re-ID ~\cite{lin_bottom-up_2019,lin_unsupervised_2020,zeng_hierarchical_2020} as illustrated Figure \ref{fig:fullyclustering}.

\paragraph{Unsupervised cross-domain person re-identification}
In Unsupervised cross-domain person Re-Id, a labelled source domain and an unlabeled target domain are used to train a Re-ID model for the target domain. The Re-ID model is pre-trained on the labelled source data in a supervised manner then adapts to the target domain with unlabelled data~\cite{fan_unsupervised_2018,fu_self-similarity_2019}. The domain knowledge of the labelled dataset is transferred to the unlabelled domain via transfer learning. In the current literature, unsupervised cross-domain person Re-ID is studied with the image dataset. A number of diverse methods  utilize transfer learning to improve unsupervised person Re-ID~\cite{lin_multi-task_2018,wang_transferable_2018,yu_unsupervised_2019,wei_person_2018,deng_image-image_2018,zhong_generalizing_2018,zhong_invariance_2019,chen_instance-guided_2019,wu_unsupervised_2019,li_cross-dataset_2019,qi_novel_2019,zhang_self-training_2019}. Some works ~\cite{lin_multi-task_2018,wang_transferable_2018} utilize extra attribute annotations to minimize the mid-level feature discrepancy. 
GAN is another popular choice for transfer learning, in which the labelled source images are transferred into images with the target image styles while preserving source identity labels. The translated images with the source identity labels are then used to fine-tune the Re-ID model for the target domain~\cite{wei_person_2018,deng_image-image_2018,zhong_generalizing_2018,zhong_invariance_2019,chen_instance-guided_2019}. Clustering-based methods has shown an advantage over the others by exploiting dynamic person features in unlabelled data. Although the knowledge transfer approach can adapt the source domain to the target domain, it assumes that the two domains have many similar features. The Domain adaptation approach may not work well when the source and the target dataset have significant feature variations.

\paragraph{Fully unsupervised person re-identification.}  The fully unsupervised Re-ID is more challenging since only unlabeled data are provided to train a Re-ID model. Thus its performance is limited without pairwise labelled data to learn invariant person features across cameras views. Therefore it is less researched as it is more challenging to learn robust representation without labelled training samples. Before deep learning-based person Re-ID, the person Re-ID models are generically using handcrafted features ~\cite{liao_person_2015,zheng_scalable_2015} and focused on metric learning to associate the query person with the hard-crafted person representation. There are few attempts of unsupervised person Re-ID with handcrafted features in which performance is far behind deep learning-based methods. Unlike unsupervised domain adaptation, fully unsupervised person Re-ID doesn't require a pre-trained model on labelled data. It fully explores the distribution of unlabeled data to learn discriminative person features. It's commonly perceived that without any label guidance it's difficulty to learn robust discriminative person features across camera views~\cite{kodirov_dictionary_2015,wang_unsupervised_2014,wang_towards_2016,yu_cross-view_2017}. A few notable achievements have greatly closed the performance gap between unsupervised cross-domain Re-ID and fully supervised Re-ID. Fully unsupervised person re-identification has been studied with both image and video person Re-ID datasets.
In reviewing papers of unsupervised cross-domain person Re-ID and fully unsupervised person Re-ID, we summarise four significant challenges that unsupervised person Re-ID solutions need to address as follows: 

\begin{itemize}
\item {\bf Lacking ground-truth identity labels to supervise feature representation learning.} A person Re-ID model is required to learn discriminative feature representation, which is trained as a classification task. Without ground-truth identity labels, the model has to predetermine the pseudo-identity labels associated with the training data, which may hinder the representation learning if the estimated identity labels are incorrect. Therefore, lacking identity labels is the first challenge to deal with for unsupervised person Re-ID.
\item {\bf Learning discriminative person feature representation with noisy pseudo-labels.} A person Re-ID model needs to learn discriminative person features from diverse person images considering occlusion, view point, illumination etc. Estimated pseudo-labels are noisy and may mislead the feature learning process. How to minimize the impact of noisy pseudo-labels and maximize the model's discriminative power is a big challenge.
\item {\bf Learning camera-invariant features without pairwise samples.} Person Re-ID is a multi-camera retrieval application. A Re-ID model is required to learn camera-invariant person features to identify the person with the cross-camera appearance in the gallery images. Without pairwise identity labels across camera views, it is challenging to learn a Re-ID model to match cross-camera person images.
\item {\bf Domain gap between datasets.} The domain gap challenge is specific to unsupervised cross-domain person Re-ID in which the person appearance show significant variation in different datasets due to difference viewpoint, light condition, background clutters etc. Therefore the domain gap needs to be considered when adapting a pre-trained model from the source to the target domain.
\end{itemize}

In the following section, we analyze unsupervised person Re-ID methods regarding the above challenges and corresponding solutions from the following aspects:
\begin{itemize}
\item {\bf Pseudo-label estimation.} Addressing the challenge of absent ground-truth identity labels by estimating pseudo-labels for unlabelled samples. The pseudo-labels are then used to supervise the model training on the unlabeled dataset. Estimated pseudo-labels are generated in an unsupervised manner and may be noisy due to false labels. We discuss how various solutions refine noisy pseudo-labels.
\item {\bf Deep feature representation learning.} Addressing the challenge of learning discriminative person feature representation with pseudo-labels from person images concerning background clutter, occlusion and poses etc.
\item {\bf Camera-invariant feature representation learning.} Addressing the challenge of learning cross-camera invariance person features. Person appearances vary in different camera views due to viewpoint, lighting condition, background etc. Commonly, an inter-class person looks similar from the same camera view, while same person appearances vary across multiple camera views. We highlight solutions that learn camera-invariant features which lead to superior performance.
\item {\bf Unsupervised domain adaptation.} Addressing the challenge of mitigating the domain gap when adapting a source model to a target domain using various unsupervised domain adaptation (UDA) strategies.
\end{itemize}

\subsection{Pseudo-Label Estimation}

The primary task of the person Re-ID problem is to learn a Re-ID model to extract discriminative person features, which minimizes the distance with the query person. As defined in Equation \ref{eq:reidobjective}, this is generally formulated as a classification task where the objective is to associate sample images with identity labels and minimize the classification losses between predicted identity labels and the ground-truth identity labels. 
 
In unsupervised person Re-ID, the ground-truth identity labels for the training samples are not provided, and the number of identities is unknown. Therefore, the first challenge in unsupervised person Re-ID is to estimate pseudo-labels for the unlabelled training samples. Apart from few early works in unsupervised cross-domain person re-identification which preserve source domain identities ~\cite{wei_person_2018,deng_image-image_2018} to train the target model, diverse solutions have been proposed to estimate identity labels based on data distribution ~\cite{fan_unsupervised_2018,ye_dynamic_2017,liu_stepwise_2017,wu_exploit_2018,zhong_invariance_2019,wu_clustering_2019}. The graph matching~\cite{ye_dynamic_2017} method proposes to use graph model to represent samples and perform dynamic graph matching for cross-camera labeling. Clustering-based methods employ clustering algorithms such as K-means and DBSCAN to progressively group training samples into clusters and use the cluster IDs as the pseudo-labels to train a Re-ID model. Clustering operation is performed progressively until the model converges and the clustering results are satisfactory~\cite{fan_unsupervised_2018,zheng_mars_2016,zhang_self-training_2019,fu_self-similarity_2019}. For example, PUL ~\cite{fan_unsupervised_2018} proposes a progressive clustering and fine-tuning method to train a Re-ID model using K-means clustering and the IDE model ~\cite{zheng_person_2016}. 

In the clustering approach, pseudo-label estimation exploits the target domain's data distribution characteristics and generates labels from clustering results. In the current literature, pseudo-labels are estimated in two flavours: hard labelling and soft labelling. A hard label is a single label predicted for a sample, such as cluster IDs in the clustering-based approach are fixed hard labels for training samples. A soft label is a probability that a label belongs to a sample. A sample could have multiple soft labels, such as in TSSL~\cite{wu_tracklet_2020} where a video tracklet can be associated with multiple identities at various probabilities. 

A common problem in clustering-based pseudo-label estimation is the label noises so that incorrect labels are predicted to guide model training. To refine the pseudo-labels, NRMT~\cite{vedaldi_unsupervised_2020} employs two networks in training to select samples and perform collaborative clustering with the selected samples and is similar to the ACT~\cite{yang_asymmetric_2020} method, which use two networks to distinguish pure and diverse samples. MEB-Net~\cite{vedaldi_multiple_2020} train multiple expert models on the source dataset and utilizes mutual learning to learn the best model from the multiple experts using a regularization term to determine the expert models' authority.

In an unlabelled dataset, the number of identities is unknown. Since the number of classes is critical in the classification task, the pseudo label estimation process needs to predetermine or estimate the number of identities in the target domain and associate identifies with the unlabeled training images. K-means is a popular clustering algorithm ~\cite{fan_unsupervised_2018} and is very sensitive to the k value. The K-means algorithm unavoidably generates false labels. Different from the K-means algorithm, BUC~\cite{lin_bottom-up_2019} uses bottom-up clustering to generate pseudo-labels. Bottom-up clustering is essentially hierarchical clustering which can visit all samples and determine the similarity of samples. Based on the similarity metric defined for cluster merging, hierarchical clustering estimate labels based on visual similarity distribution. Hard samples such as the visually similar person with different identities can't be distinguished in hierarchical clustering, thus merge hard samples into the same cluster. These false labels eventually mislead the model training and hamper the overall Re-ID performance. To better separate the positive samples from negative samples, GDS~\cite{vedaldi_global_2020} employs a global distance-distribution constraint to separate positive and negative samples, therefore achieve better pseudo-labels.  

To reduce the impact of the hard samples, HCT~\cite{zeng_hierarchical_2020} uses PK (K samples out of P identities) sampling to form a new dataset and employs a hard-batch triplet loss to similar samples closer and separate different samples further from each other, therefore reduce the influence of hard samples. DSCE\cite{yang_joint_2021} uses DBSCAN for pseudo-label generation and doesn't deal with hard samples, which pose greater feature variation and could be used to learn invariant features across scenes. SSG \cite{fu_self-similarity_2019} builds clusters from different camera views for similar global bodies and local parts. The cluster IDs are then used to supervise the model training. MAR~\cite{yu_unsupervised_2019} addresses the hard samples by using comparative consistency between soft labels. It learns soft labels from the source dataset for Re-ID model training to reduce the impact of the hard samples. MMT~\cite{ge_mutual_2020} refine hard labels offline and soft labels online in an alternative training manner. 

BUC~\cite{lin_bottom-up_2019} and HTC~\cite{zeng_hierarchical_2020} are essentially hierarchical clustering algorithms. The hierarchical structure indicates that the distance criteria used for cluster merging or division are crucial. In hierarchical clustering, clustering and cluster merging depends on the distance criteria. BUC exploit a minimum distance criterion to merge clusters by taking the shortest distance between images so that if two images are very similar, their clusters will be merged. BUC introduces a diversity regularisation term into the distance criterion to boost diversity and avoid dominant clusters as express in Equation \ref{eq:minimus}. \begin{equation} \label{eq:minimus}
D(A,B) = \min_{x_a \in A, x_b \in B} d(x_a, x_b) + \lambda(|A| + |B|).
\end{equation}
where $d(x_a, x_b)$ is the Euclidean distance between the two images. The minimum criteria work well to merge visually similar samples into the same cluster. However, it doesn't use other samples and fails to consider outliers in the sample space, such as hard samples that are similar in appearance but with different identities.

To reduce the impact of outliers and false labels in the clustering process,  HCT uses Euclidean distance to measure the sample distance, which can be expressed as: 
\begin{equation} \label{eq:htc}
D_{ab} = -\frac{1}{n_an_b}\sum_{i \in C_a,j \in C_b} D(C_{a_i},C_{b_j}).
\end{equation}
where $C_{a_i}$ ,$C_{b_j}$ are two samples in the cluster $C_a$ and $C_b$ respectively. $n_a$ and $n_b$ are sample numbers in $C_a$ and $C_b$. $D(\cdot)$ is the Euclidean distance. It considers all the pairwise distances between clusters. All pairwise distances are weighted equally. Therefore, it effectively alleviate the impact of the hard samples, resulting in better clustering results.

To refine pseudo-labels in unsupervised video-based person Re-ID, SMP~\cite{liu_stepwise_2017} uses a step-wise metric promotion method to predict tracklet labels progressively. SMP method uses reciprocal nearest neighbour search to eliminate the hard samples which generate false labels. An OIM ~\cite{xiao_joint_2017} loss function instead of the cross-entropy loss is used in the classification operation to speed up model convergence with a large number of identities.

\subsection{Deep Feature Representation Learning}
The core part of the Re-ID system is to learn a Re-ID model to extract discriminative person feature representation, which minimizes the distance with the query person (Eq.\ref{eq:reidobjective}). With estimated pseudo-labels $y_i$, unsupervised person Re-Id models are trained in a supervised manner to learn discriminative feature representation. However, due to the noisy pseudo-labels, the feature learning process is biased inevitably. The general approach to alleviating this impact is to exploit the training data distributions rather than solely relying on the pseudo-labels to guide the feature learning~\cite{wang_unsupervised_2020,xiang_second-order_2020,zhai_ad-cluster_2020,zeng_hierarchical_2020}.

Deep feature representation learning in unsupervised models exploit global and local person features similar to supervised person Re-ID models. Global feature learning, such as image-level features, has been well studies traditionally. Discriminative local features have been proved effective in supervised person Re-ID ~\cite{zhao_deeply-learned_2017,xiang_part-aware_2020,guo_beyond_2019}. For Unsupervised person Re-ID,  PAUL~\cite{yang_patch-based_2019} proposes to learn local discriminative features from unlabeled patches rather than from the global images in an unsupervised manner. ADTC~\cite{vedaldi_attention-driven_2020} employs an attention mechanism to strengthen the informative parts of person images. SSG~\cite{fu_self-similarity_2019} mines the visual similarities from the global body level as well as the local level from upper and lower body parts. PAST~\cite{zhang_self-training_2019} exploits local features of target data with triplet losses, leading to improved feature representations. Unlike the PAST methods, which use fixed samples, DCML~\cite{vedaldi_deep_2020} emphasizes credible samples for feature representation learning to reduce the impact of the false labels. The DCML method proposes two metrics, the KNN similarity and the prototype similarity. The KNN similarity measures the neighbourhood density between the sample and its neighbours. The prototype similarity estimates the sample and the class prototype's similarity. The DCML discards hard samples that provide more discriminate information across cameras.  

Efficient loss functions play critical roles in guiding deep feature learning. The general purpose is to close the intra-class samples in the feature space and drive them away from the inter-class samples. Euclidean distance and cosine similarity are common metrics to evaluate the similarity level among samples. The Re-ID model is generally formulated as a classification problem where conventional cross-entropy loss trains the classifier. BUC~\cite{lin_bottom-up_2019} uses a repelled loss to maximize the sample balance in identities. To close intra-class features while separating them from other classes, the triplet loss ~\cite{hermans_defense_2017} is a popular choice in guiding feature learning for a large number of classes ~\cite{wang_unsupervised_2020,xiang_second-order_2020,zhai_ad-cluster_2020}. The triplet loss is formulated as:
\begin{equation} \label{eq:triplet}
L_{tri} = \sum_{i=1}^P \left[m+D(x_a^i,x_p^j)-D(x_a^i,x_n^j)\right].
\end{equation}
where $x_a^i$ a is the anchor, $x_p^j$ is the person features with the same identity as the anchor. $x_n^j$ is the person feature with different identity as the anchor. $D(\cdot)$ is the Euclidean distance, and $m$ is the hyper-parameter margin. This loss ensures that intra-class features are close in feature space while away from the inter-class features. To reducing the impact of outliers in unsupervised clustering, HCT~\cite{zeng_hierarchical_2020} uses a hard-batch triplet loss (Eq.\ref{eq:hard-batch}) to guide feature learning.
\small
\begin{equation} \label{eq:hard-batch}
L_{htri}=\sum_{i=1}^P \sum_{a=1}^K \left[m+\max_{p=1...K}D(x_a^i,x_p^j)-\min_{\substack{j=1...P\\ n=1...N}}D(x_a^i,x_n^j) \right].
\end{equation}
\normalsize

Hard-batch triplet loss ensures that give an anchor $x_a^i$, $x_p^j$ is closer to $x_a^i$ than $x_n^j$. As a result, a person with the same identity will be closer to each other and separate from others.

Unsupervised person Re-ID is also applicable in video-based person Re-ID applications, where a video sequence with multiple frames (tracklet) represents a person. Due to the rich appearance and temporal information, video-based unsupervised Re-ID has increased interest in the computer vision community, bringing additional challenges in video feature representation learning with tracklets. In video-based person Re-ID, temporal information is widely used in feature learning. JVTC~\cite{vedaldi_joint_2020-1} computes the temporal consistency based on the distribution of time interval between cameras. The intuition is easy to follow. For example, when a person appears in camera $A$ at time $t_a$, according to the temporal distribution, the person would be likely recorded by camera $B$ at time $t_b$, and is less likely be recorded by camera $C$ at time $t_c$. This temporal information is useful to determine visually similar hard samples, thus improve representation learning. 

Tracklets are useful in learning invariant feature representation simply because that same tracklet frames with various person appearances represent the same person. Person  tracklets can be obtained by tracking algorithms~\cite{leal-taixe_motchallenge_2015,ristani_performance_2016,gao_graph_2019,gao_i_2019,zhang_deeper_2019} without prohibitive labelling cost. TAUDL~\cite{ferrari_unsupervised_2018} and UTAL~\cite{li_unsupervised_2020} learns intra-camera person tracklet features and correlate inter-camera tracklets to learn better person features. TSSL~\cite{wu_tracklet_2020} proposes a self-supervised learning method that directly learn features from unlabelled tracklet data for both video-based and image-based unsupervised person Re-ID. In addition, TSSL predicts soft labels for images to minimize the impact of hard samples. 

\subsection{Camera-Aware Invariance Learning}

Person Re-ID is a cross-camera retrieval process aiming to learn a model to discriminate samples from different camera views. Suppose a model trained with samples from some cameras can also generalize to distinguish samples from the other cameras. In that case, we could obtain a model that can extract the intrinsic feature without camera-specific bias and is robust to camera changes. However, due to lacking pairwise ground-truth information for cross-camera samples, unsupervised Re-ID models ability to learn camera-aware invariant features is limited.  

To learn camera-aware invariant features, SCCT~\cite{xiang_second-order_2020} calculates the colour statistics of different camera views and apply a linear transformation to match the statistics of transformed images same as target images, thus improves the camera-aware invariance. Cross-domain Mixup~\cite{vedaldi_generalizing_2020} conducts interpolation on the data manifold, which is similar to GAN-based image style transfer, and achieve good cross-camera invariant features. DSCE\cite{yang_joint_2021} introduces a camera-aware meta-learning (MetaCam) aiming to learn camera-invariant representations by simulating the cross-camera Re-ID process during training. Precisely, MetaCam separates the training data into meta-train and meta-test subsets, ensuring that they belong to entirely different cameras. Then enforce the model to learn camera-invariant features under both camera settings by updating the model with meta-train and validating the updated model with meta-test. UGA~\cite{wu_unsupervised_2019-1} mines the cross-camera association and alleviates the impact of noisy labels. Specifically, UGA trains the model in two stages. The intra-camera learning stage is to learn person representations considering the camera information, which helps to reduce false cross-camera relationships in the inter-camera learning stage. UCDA~\cite{qi_novel_2019} utilizes the unique characteristics of camera-level images and learn camera-aware invariant features.

In unsupervised clustering, each sample is allocated a label the same as its nearest neighbours. Utilizing a memory bank (e.g. lookup table), a sample's probability sharing the same identity with its neighbours can be calculated as: 
\begin{equation} \label{eq:oimprob}
p_{ij} = \frac{\exp(s.m_j^Tx_i^t)}{\sum_{k=1}^{N_t}\exp(s.m_k^Tx_i^t)}.
\end{equation}
where $s$ is a scaling factor $m_i^T$ is the person features for the $i_{th}$ target identity in the lookup table (LUT). $m_j^T$ is the $j_{th}$ person features in the LUT. The objective is to minimise the loss:
\begin{equation} \label{eq:oimloss}
\mathcal{L} = -\log(p_t). 
\end{equation}

Due to the cross-camera scene variation, the different intra-camera people may share more significant similarities than those in the inter-camera scene. HHL~\cite{zhong_generalizing_2018} learns camera-aware person features in the unlabeled domain. However, HHL overfits the visually similar pairs from the target domain. Consequently, the Re-ID model is sensitive to pose and background variations in the target domain. To alleviate the impact of the hard samples, the Mixup~\cite{vedaldi_generalizing_2020} method considers the camera information when calculating the probability that a sample feature shares the same identity as another feature. The probability that two features of the same identify are from the same camera view can be expressed as:
\begin{equation} \label{eq:intra}
p_{ij}^{intra} = \frac{\exp(s.m_j^Tx_i^t)}{\sum_{k \in O_i^{intra}}\exp(s.m_k^Tx_i^t)}.
\end{equation}

Similarly, the probability that two samples of the same identity are from different camera views is formulated as:
\begin{equation} \label{eq:inter}
p_{ij}^{inter} = \frac{\exp(s.m_j^Tx_i^t)}{\sum_{k \in O_i^{inter}}\exp(s.m_k^Tx_i^t)}.
\end{equation}
where $s$ is a scaling factor, $m_i^T$ is the person features for the $i_{th}$ target identity in the lookup table (LUT). $m_j^T$ is the $j_{th}$ person features in the LUT. The objective is to minimise the the combination of intra-camera and inter-camera losses:
\begin{equation} \label{eq:oimloss}
\mathcal{L} = -\sum w_{i,j}\log(p_{i,j}^{intra})-\sum w_{i,t}\log(p_{i,t}^{inter}). 
\end{equation}
where $j$ denotes the $j_{th}$ intra-camera neighbor of $x_i^t$, and $t$ denotes the $t_{th}$ inter-camera neighbor of $x_i^t$. By taking consideration of inter-camera neighbor in calculating identity probability, the learnt feature achieve camera-aware invariance.

To learn camera-aware invariant feature in video-based unsupervised person Re-ID, TAUDL~\cite{ferrari_unsupervised_2018} learns intra-camera person tracklet features first, then jointly learn all the intra-camera tracklet features to propagate feature learning from intra-camera learning to inter-camera feature learning, therefore achieving camera-aware invariance. 

\subsection{Unsupervised Domain Adaption}\label{uda}

Unsupervised domain adaptation addresses the domain gap challenge in unsupervised cross-domain person Re-ID. In this setting, a model is first pre-trained on the labelled source domain and then adapted to the unlabelled target domain. In unsupervised cross-domain person Re-ID, there is an unlabeled target domain $T = \{t_i\}^{N^T}_{i=1}$ containing $N^T$ person images. Additionally, a labelled source domain $S = \{si, yi\}^{N^S}_{i=1}$ containing $N^S$ labelled person images is available to pre-train the Re-ID model, where $y_i$ is the ground-truth identity label for the training sample $s_i$. The objective of the cross-domain person Re-ID is to learn a feature extractor $f(t)$ for the unlabeled target domain $T$, using both source domain $S$ and target domain $T$. The training cost of $f(t)$ is the sum of the training loss on both pre-trained and adapted models. With ground-truth identity labels, the training on source domain $S$ is a classification task with the conventional cross-entropy loss, i.e.,
\begin{equation} \label{eq:eq4}
\mathcal{L_S} = -\frac{1}{N_S}\sum_{i=1}^{N_S} \log P(y_i|s_i),
\end{equation}
where $P(y_i|s_i)$ is the probability of sample $s_i$ being identity $y_i$.
This supervised learning on source domain ensures the optimal performance of $f(t)$ on source domain thus ensure the feature extraction quality on the target domain for adaptation.

Ground-truth identity labels are absent in training the the target domain, instead, pseudo-labels are used to supervise the training. The training loss can be expressed as:
\begin{equation} \label{eq:eq4}
\mathcal{L_T} = -\frac{1}{N_T}\sum_{i=1}^{N_T} \log P(\hat{y_i}|t_i),
\end{equation}
where $\hat{y_i}$ is the estimated pseudo-label for $t_i$.  $P(\hat{y_i}|t_i)$ is the probability of sample $t_i$ being identity $\hat{y_i}$. 

The pre-trained Re-ID model on the source domain is sensitive to the feature variations in the target domain. The domain gap has a high impact on the cross-domain Re-Id performance. Therefore, the image variations must be considered when adapting the pre-trained model to the target domain. The existing unsupervised domain adaptation solutions can be grouped into three categories, mid-level feature alignment, image style transfer and clustering-based approach.
\begin{figure*}[t]
\begin{center}
\includegraphics [width=.75\linewidth]{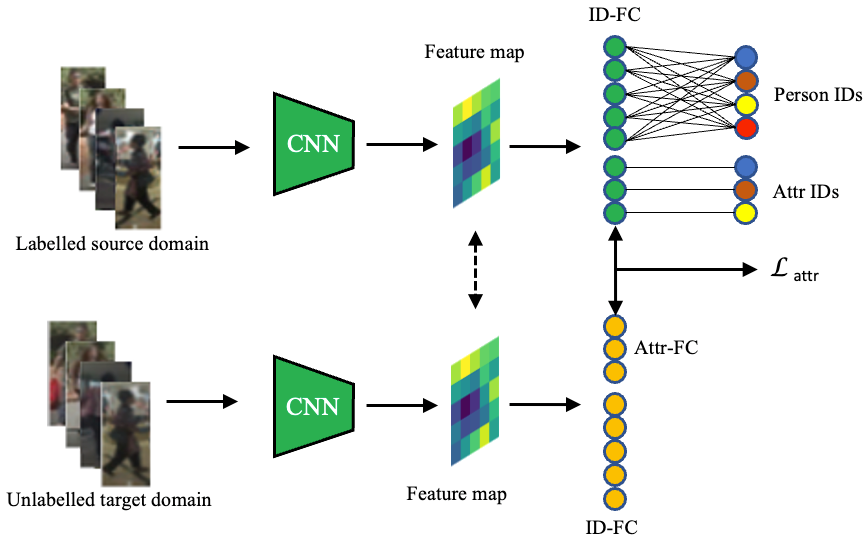}
\end{center}
  \caption{ Illustration of feature alignment approach for unsupervised domain adaptation. Mid-level attribute features are aligned between source and target domains in a joint learning pipeline. $\mathcal{L}_{attr}$ represents the attribute alignment loss between the source attributes and the target attributes.}
\label{fig:fa}
\end{figure*}
\subsubsection{Mid-Level Feature Alignment}
The feature alignment approach aims to reduce the domain gap at the feature and image levels and assumes that the source and target datasets share a common mid-level feature space, and the common mid-level features can be used to infer person identities cross domain~\cite{lin_multi-task_2018,wang_transferable_2018}. ~\cite{wang_transferable_2018,zhong_generalizing_2018,lv_unsupervised_2018} exploit auxiliary information to improve the model generalization capability. TFusion~\cite{lv_unsupervised_2018} exploits Spatio-temporal information to learn better feature representation. TJ-AIDL~\cite{wang_transferable_2018} and MMFA~\cite{lin_multi-task_2018} utilize annotated attributes information to learn discriminative feature representation. An example of the attribute alignment approach is illustrated in Figure \ref{fig:fa}. However, strictly speaking, annotated attributes are considered supervised learning and incur labelling costs. D-MMD~\cite{vedaldi_unsupervised_2020-1} loss is proposed to align features by closing pairwise distances using small batch sizes. In PAUL~\cite{yang_patch-based_2019}, it is assumed that if two images are similar, then their local patches are similar. So that instead of learning the image level features, PAUL resorts to patch features for unsupervised person Re-ID. Differ from the above direct feature alignment methods, $p$MR-SADA~\cite{wang_smoothing_2020} proposes to reduce the domain gap at the image level by mixing the camera semantic information with the original images.

To alleviate the negative transfer caused by the domain divergence,  DAAM~\cite{huang_domain_2020} proposes an attention model based on residual mechanism. It transfers knowledge from the labelled dataset to the unlabeled dataset by jointly modelling the domain-shared and domain-specific features. Moreover, it differs significantly from existing methods in that a soft label loss is proposed to alleviate the negative effect of inaccuracy pseudo labels.

\subsubsection{Image Style Transfer}

\begin{figure*}[t]
\begin{center}
\includegraphics [width=0.8\linewidth]{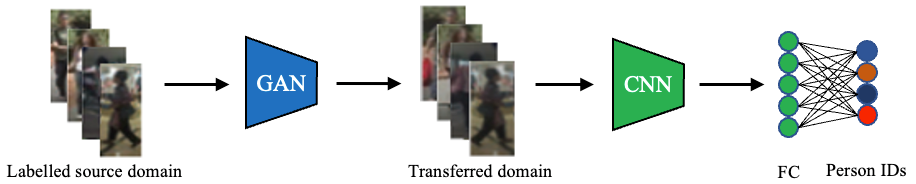}
\end{center}
  \caption{Illustration of GAN-based image transfer for unsupervised domain adaptation. Source images are transferred to target image styles using a GAN while preserving source identity labels. The transferred images with the source identity labels are used to train the Re-Id model as a classification task.}
\label{fig:gan}
\end{figure*}

Image style transfer, in particular, GAN-based image style transfer has been an popular approach to transfer knowledge between the source domain and the target domain for unsupervised cross-domain person Re-ID ~\cite{li_cross-dataset_2019,liu_adaptive_2019,zhong_generalizing_2018,wei_person_2018,deng_image-image_2018,liu_adaptive_2019}. For example, PTGAN ~\cite{wei_person_2018} and SPGAN~\cite{deng_image-image_2018} directly use CycleGAN~\cite{zhu_unpaired_2017} to reduce the domain gap problem. They first transfer images from source datasets to the target style while preserving source identity labels, then uses transferred images and preserved source domain identity labels to train a Re-ID model for the target domain. An example of GAN-based image transfer approach is illustrated in Figure \ref{fig:gan}. HHL ~\cite{zhong_generalizing_2018} generates images under different cameras and trains networks using triplet loss, thus improves model performance with the camera-aware invariant features. ECN ~\cite{zhong_invariance_2019} utilizes transfer learning and minimizes the target invariance using an exemplar memory to learning invariant features. CR-GAN\cite{chen_instance-guided_2019} exploits contextual styles, in particular the background. It masks the person in target images to retain background clutter and use the source person and the target background as transferred images to train the Re-ID model. DAS~\cite{bak_domain_2018} employs a new synthetic dataset simulating various lighting conditions and transfer the synthetic dataset to the target styles using CycleGAN~\cite{zhu_unpaired_2017}. PDA-Net~\cite{li_cross-dataset_2019} exploits PatchGAN~\cite{isola_image--image_2017} and poses disentanglement in transferring images from source to target. Based on SPGAN ~\cite{deng_image-image_2018}, CR-GAN~\cite{chen_instance-guided_2019} PTGAN~\cite{wei_person_2018} use person segmentation masks as extra information to preserve the discriminative person features. 

\subsubsection{Clustering-Based Approach}

Clustering-based methods adapt the pre-trained model in the source domain in two steps. First, they use the pre-trained model to extract and estimate pseudo-labels based on unlabeled image cluster IDs. Then they use the images with pseudo-labels to fine-tune the pre-trained model on the target dataset. The pseudo-labels are generated by clustering all cross-camera samples based on visual similarity. The model is then fine-tuned as a classification task. These two stages are executed alternately to optimize the Re-ID performance with intra-camera and inter-camera losses. An example of clustering-based UDA is illustrated in Figure \ref{fig:clustering}. Since the Re-ID model is trained as a classification task, one key parameter in the clustering-based approach is to determine the number of identities for the final classification output. CDS~\cite{wu_clustering_2019} adopts the K-means algorithm to cluster target samples, where K is pre-defined empirically. CDS uses a dynamic sampling strategy to determine whether a sample belongs to a cluster by calculating the cosine similarity between the sample and the cluster centroid. 

Clustering results based on sample visual similarity lead to data imbalance in clusters. To deal with data imbalance, DBC~\cite{ding_towards_2019} learns the data distribution then uses pairwise sample relationships to achieve better cluster balance in the clustering process. MMCL~\cite{wang_unsupervised_2020} formulates a multi-label classification task to progressively estimate soft pseudo-labels. The predicted labels are verified based on visual similarity and cycle consistency. AD-Cluster\cite{zhai_ad-cluster_2020} employs an image generator to augment samples in the density-based clustering and improve the discriminative capability with the augmented clusters.

In unsupervised video-based person Re-Id, temporal information can be easily obtained in a multi-camera system. JVTC~\cite{vedaldi_joint_2020-1} tackles the domain gap challenge by enforcing visual and temporal consistency in a multi-class classification task. Visual similarity and temporal consistency are used in multi-class prediction to optimize the intra-class and inter-class association. Two samples are considered the same identity only when they share considerable visual similarity and satisfy the temporal consistency. To accurately estimate labels from noisy frames, RACE ~\cite{ye_robust_2018} proposes robust anchor embedding and top-$k$ counts label prediction to learn full feature embeddings. SSL~\cite{lin_unsupervised_2020} offers a soft label classification approach to exploit the unlabeled image similarity fully. DAL~\cite{chen_deep_2018} proposes two margin-based association losses to effectively constrain the intra-camera and inter-camera frame association, thus achieve better discriminative cross-camera person representation. ACT ~\cite{yang_asymmetric_2020} design an asymmetric co-teaching framework with two models to select samples from each other. One model focus on training with true positive sample and another model address the hard samples with false-positive samples.

\section{Datasets and Evaluation}\label{evaluation}
\subsection{Datasets}
Unsupervised person Re-ID aims to address the data scalability issue in conventional supervised person Re-ID. Most models are trained and evaluated on large-scale person Re-ID datasets sourced from the multi-camera video system. Person Re-ID is generally perceived as a multi-camera application. Therefore, the person appearing in more than one camera views are used to train and test a person Re-ID model. Dataset statistics are summarised in Table \ref{table:datasets}.   

\paragraph{CUHK03~\cite{li_deepreid_2014}} has 1,360 identities in 13,164 images from 6 camera views. This dataset has both manually annotated bounding boxes and auto-detected bounding boxes. 1,160 identities are used for training, 200 identities are selected for testing.

\paragraph{Market-1501~\cite{zheng_scalable_2015}} is a large-scale benchmark dataset for image-based person Re-ID. It has 1,501 identities and 32,668 annotated person using the DPM ~\cite{girshick_deformable_2014} detector. On average, each person has 3.6 images from each camera. Among the 1501 identities, 750 identities are used for training, and the rest 751 identities are used for evaluation. The evaluation set has 19,732 images. 3,368 images are used as probe queries.

\paragraph{DukeMTMC-Re-ID~\cite{zheng_unlabeled_2017}} is extracted from the Duke-MTMC~\cite{ristani_performance_2016} tracking dataset for image-based person Re-ID. The person images are manually cropped from high-resolution videos from 8 non-overlapping cameras. It has 34,183 person images and 1,404 identities that appear in at least two cameras. The training set has 16,522 images of 702 identities. 17,661 images and 702 identities are used for evaluation. In evaluation, 2,228 images are used as queries.
\paragraph{MSMT17~\cite{wei_person_2018}} is the largest dataset for image-based person Re-ID. It contains videos captured by 15 cameras at 12 time intervals. The videos are 180 hours in length and pose complex photometric variations. For image-based person Re-ID, the dataset has 4,101 identities and 126,441 person images.

\begin{table} [t]
\begin{center}
\caption{Commonly used benchmark datasets for unsupervised person re-identification. The top four datasets are for image-based person Re-ID, the bottom four are for video-based person Re-ID.}
\label{table:datasets}
\resizebox{\columnwidth}{!}{
\begin{tabular}{l|rrrccc}
\toprule
{\bf Dataset} & \#boxes & \#identity & \#cameras & detector \\
\midrule
CUHK03~\cite{li_deepreid_2014} & 13,164 &1,360&6 &hand/detector \\
Market-1501~\cite{zheng_scalable_2015} &32,668 &1,501 & 6 & hand \\
DukeMTMC-Re-ID~\cite{zheng_unlabeled_2017} &34,183 &1,404& 8& hand \\
MSMT17~\cite{wei_person_2018} & 126,441 & 4,101 & 15 & Faster R-CNN \\
\midrule
PRID2011~\cite{hirzer_person_2011}&40,033&200&2&hand\\
MARS ~\cite{zheng_scalable_2015}& 1,067,516 & 1,261& 6 & DPM \\
DukeMTMC-VideoRe-ID~\cite{wu_exploit_2018} & 815,420&1,404&8&hand\\
iLIDS-VID~\cite{wang_person_2014}&42,460 & 300 & 2 & hand\\
\bottomrule
\end{tabular}
}

\label{table:image-datasets}
\end{center}
\end{table}

\paragraph{PRID2011~\cite{hirzer_person_2011}} is an early video dataset that contains person videos from only two cameras. The first camera captured 385 person identities, 749 person identities were from the second camera. 200 person appeared in both cameras. The videos are curated to tracklets (sequence of frames) with 5 to 675 frames. 

\paragraph{MARS~\cite{zheng_scalable_2015}} is a large-scale dataset for video-based person Re-ID. Videos were captured by 6 cameras. 1,261 person identities appeared in at least 2 cameras. Lighting conditions and image quality vary in the dataset. To mimic the real-life viewing condition, the dataset contains 3,248 distractors. Tracklets are generated by DPM and GMMCP ~\cite{dehghan_gmmcp_2015} tracker.

\paragraph{DukeMTMC-VideoRe-ID~\cite{wu_exploit_2018}}  is a subset of the Duke-MTMC~\cite{ristani_performance_2016} tracking dataset for video-based person Re-ID. It contains 1,812 identities and 4,832 tracklets from 8 cameras. On average, a tracklet has 168 consecutive frames. The person boxes in the frames are manually cropped and labelled.

\paragraph{iLIDS-VID~\cite{wang_person_2014}} is for video-based person Re-ID. It contains 300 person identities captured by 2 cameras. 600 tracklets are hand-annotated, covering 300 person identities from 2 camera views. Tracklet lengths vary from 23 to 192 frames.

\subsection{Evaluation Metrics}
Person Re-ID methods are commonly evaluated with mean averaged precision (mAP) and cumulative matching characteristics (CMC Rank-K). Inherited from the object detection problem, the mAP is a popular evaluation metric in person Re-ID. With mAP, an averaged precision (AP) is calculated for each probe image, and then the final evaluation mAP is the average of all APs. With CMC, the top K similar person in the gallery images are ranked according to the intersection-over-union (IoU@0.5) overlap with the ground-truth person identity. Most evaluation protocols using single person image as the query person.
\begin{table*} [t]
\caption{Image-based unsupervised cross-domain person Re-ID evaluation results. Adaptation approach is annotated as described in section \ref{uda}. FA: mid-level feature alignment approach. IT: image style transfer approach. Cluster: clustering-based approach.}
\label{table:uda}
\resizebox{\tblwidth}{!}{
\begin{tabular}{l|l|l|rrrr|rrrr|rrrr|rrrr}
\multicolumn{1}{c}{}&\multicolumn{1}{c}{}&\multicolumn{1}{c}{}& \multicolumn{4}{c}{{\bf Duke $\rightarrow$ Market-1501}} & \multicolumn{4}{c}{{\bf Market-1501 $\rightarrow$ Duke}} & \multicolumn{4}{c}{{\bf Market-1501 $\rightarrow$ MSMT17}} & \multicolumn{4}{c}{{\bf Duke $\rightarrow$ MSMT17}} \\
\toprule
{\bf Method} &Year&Approach& R-1 & R-5 & R-10& mAP & R-1 & R-5 & R-10 & mAP& R-1 & R-5 & R-10 & mAP& R-1 & R-5 & R-10 & mAP \\
\midrule
CAMEL~\cite{yu_cross-view_2017}&2017&Cluster&54.50&-&-&26.30&-&-&-&-&-&-&-&-&-&-&-&-\\
PUL~\cite{fan_unsupervised_2018}&2018&Cluster&44.70&59.10&65.60&20.10&30.40&44.50&50.70&16.40&-&-&-&-&-&-&-&-\\
MMFA~\cite{lin_multi-task_2018}&2018&FA&56.70&75.00&81.80&27.40&45.30&59.80&66.30&24.70&-&-&-&-&-&-&-&-\\
TJ-AIDL~\cite{wang_transferable_2018}&2018&FA&58.20&74.80&81.10&26.50&44.30&59.60&65.00&23.00&-&-&-&-&-&-&-&-\\
HHL (g) ~\cite{zhong_generalizing_2018}&2018&IT&62.20&78.80&84.00&31.40&46.90&61.00&66.70&27.20&-&-&-&-&-&-&-&-\\
PTGAN~\cite{wei_person_2018}&2018&IT&38.60&-&66.10&-&27.40&-&50.70&-&10.20&-&24.40&2.90&11.80&-&27.40&3.30\\
SPGAN ~\cite{deng_image-image_2018}&2018&IT&58.10&76.00&82.70&26.90&46.90&62.60&68.50&26.40&-&-&-&-&-&-&-&-\\
DAS* ~\cite{bak_domain_2018}&2018&IT& 65.70 &-&-&-&-&-&-&-&-&-&-&-&-&-&-&-\\
DAR (c)~\cite{song_unsupervised_2018}&2018&Cluster&75.80&89.50&93.20&53.70&68.40&80.10&83.50&49.00&-&-&-&-&-&-&-&-\\
PAUL~\cite{yang_patch-based_2019}&2019&FA&68.50&82.40&87.40&40.10&72.00&82.70&86.00&53.20&-&-&-&-&-&-&-&-\\
CDS~\cite{wu_clustering_2019}&2019&Cluster&71.60&81.20&84.70&39.90&67.20&75.90&79.40&42.70&-&-&-&-&-&-&-&-\\
CamStyle~\cite{zhong_camstyle_2019}&2019&IT&64.70&80.20&85.30&30.40&51.70&67.00&72.80&27.70&-&-&-&-&-&-&-&-\\
ATNet \cite{liu_adaptive_2019}&2019&IT&55.70&73.20&79.40&25.60&45.10&59.50&64.20&24.90&-&-&-&-&-&-&-&-\\
CR-GAN~\cite{chen_instance-guided_2019}&2019&IT&77.70&89.70&92.70&54.00&68.90&80.20&84.70&48.60&-&-&-&-&-&-&-&-\\
CASCL~\cite{wu_unsupervised_2019}&2019&FA&64.70&80.20&85.60&35.60&51.50&66.70&71.70&30.50&-&-&-&-&-&-&-&-\\
PDA-Net~\cite{li_cross-dataset_2019}&2019&IT&75.20&86.30&90.20&47.60&63.20&77.00&82.50&45.10&-&-&-&-&-&-&-&-\\
UCDA~\cite{qi_novel_2019}&2019&Cluster&64.30&-&-&34.50&55.40&-&-&36.70&-&-&-&-&-&-&-&-\\
MAR~\cite{yu_unsupervised_2019}&2019&Cluster&67.70&81.90&-&40.00&67.10&79.80&-&48.00&-&-&-&-&-&-&-&-\\
ECN~\cite{zhong_invariance_2019}&2019&Cluster&75.10&87.60&91.60&43.00&63.30&75.80&80.40&40.40&25.30&36.30&42.10&8.50&30.20&41.50&46.80&10.20\\
SSG (c)~\cite{fu_self-similarity_2019}&2019&Cluster&86.20&94.60&96.50&68.70&76.00&85.80&89.30&60.30&31.60&-&49.60&13.20&32.20&-&51.20&13.30\\
DG-Net++~\cite{zheng_joint_2019}& 2019&IT& 82.10 & 90.20 & 92.70 & 61.70 & 78.90 & 87.80 & 90.40 & 63.80 & 48.40 & 60.90 & 66.10 & 22.10 & 48.80 & 60.90 & 65.90 & 22.10 \\
PAST~\cite{zhang_self-training_2019}&2019&Cluster&78.38&-&-&54.62&72.35&-&-&54.26&-&-&-&-&-&-&-&-\\
DCML~\cite{vedaldi_deep_2020}&2020&Cluster&88.20&94.90&96.40&72.30&79.30&86.70&89.50&63.5&-&-&-&-&-&-&-&-\\
D-MMD~\cite{vedaldi_unsupervised_2020-1}&2020&FA&70.60&87.00&91.50&48.80&63.50&78.80&83.90&46.00&29.10&46.30&54.10&13.50&34.40&51.10&58.50&15.30\\
ADTC~\cite{vedaldi_attention-driven_2020}&2020&Cluster&79.30&90.80&94.10&59.70&71.90&84.10&87.50&52.50&-&-&-&-&-&-&-&-\\
ACT~\cite{yang_asymmetric_2020}&2020&Cluster&80.50&-&-&60.60&72.40&-&-&54.50&-&-&-&-&-&-&-&-\\
pMR-SADA~\cite{wang_smoothing_2020}&2020&FA&83.00&91.80&94.10&59.80&74.50&85.30&88.70&55.80&-&-&-&-&-&-&-&-\\
DAAM~\cite{huang_domain_2020}&2020&FA&86.40&-&-&67.80&77.60&-&-&63.90&44.50&-&-&20.80&46.70&-&-&21.60\\
MMCL~\cite{wang_unsupervised_2020}&2020&Cluster&84.40&92.80&95.00&60.40&72.40&82.90&85.00&51.40&-&-&-&-&43.60&54.30&58.90&16.20\\
SCCT~\cite{xiang_second-order_2020}&2020&Cluster&86.60&94.50&96.90&67.90&76.00&85.00&88.90&60.40&23.60&36.00&42.10&8.30&37.80&51.20&57.00&13.20\\
AD-Cluster~\cite{zhai_ad-cluster_2020}&2020&Cluster&86.70&94.40&96.50&68.30&72.60&82.50&85.50&54.10&-&-&-&-&-&-&-&-\\
JVTC~\cite{vedaldi_joint_2020-1}&2020&Cluster&86.80&95.20&97.10&67.20&80.40&89.90&92.20&66.50&48.60&{\bf 65.30}&68.20&25.10&52.90&70.50&75.90&27.50\\
MMT~\cite{ge_mutual_2020}&2020&Cluster&87.70&94.90&96.90&71.20&78.00&88.80&92.50&63.10&49.20&63.10&68.80&22.90&50.10&63.90&69.80&23.30\\
NRMT~\cite{vedaldi_unsupervised_2020}&2020&Cluster&87.80&94.60&96.50&71.70&77.80&86.90&89.50&62.20&-&-&-&-&-&-&-&-\\
Mixup~\cite{vedaldi_generalizing_2020}&2020&IT&88.10&94.40&96.20&79.50&79.50&88.30&91.40&65.20&43.70&56.10&61.90&20.40&51.70&64.00&68.90&24.30\\
GDS~\cite{vedaldi_global_2020}&2020&Cluster&89.30&-&-&72.50&76.70&-&-&59.70&-&-&-&-&-&-&-&-\\
MEB-Net~\cite{vedaldi_multiple_2020}&2020&Cluster&89.90&96.00&97.50&76.00&79.60&88.30&92.20&66.10&-&-&-&-&-&-&-&-\\
SPCL~\cite{ge_self-paced_2020}&2020&Cluster&-&-&-&-&-&-&-&-&{\bf 53.70}&65.00&{\bf 69.80}&{\bf 26.80}&-&-&-&-\\
DSCE\cite{yang_joint_2021}&2021&Cluster&90.10&-&-&76.50&79.50&-&-&65.00&-&-&-&-&-&-&-&-\\
GLT~\cite{zheng_group-aware_2021}&2021&Cluster&{\bf 92.20}&{\bf 96.50}&{\bf 97.80}&{\bf 79.50}&{\bf 82.00}&{\bf 90.20}&{\bf 92.80}&{\bf 69.20}&-&-&-&-&-&-&-&-\\

\bottomrule
\end{tabular}
}
\end{table*}
\subsection{Performance Analysis}

This section summarises and analyses the evaluation results considering the significant challenges in unsupervised person Re-ID. We aim to present the influential factors that contribute to the overall performance of relevant models. In person Re-ID, the pre-trained ResNet on ImageNet is wildly used as the initial model and updated with various objectives. ImageNet has many person images, which make the pre-trained ResNet model suitable for initial person feature extraction. Therefore, in this survey, we don't specifically discuss CNN backbones and network architectures. Instead, we highlight how the innovative solutions address the underlying challenges.

As discussed in Section \ref{main}, unsupervised person Re-ID has two main settings, unsupervised cross-domain person Re-ID and fully unsupervised person Re-ID.  Unsupervised cross-domain person Re-ID has been solely investigated in image-based person Re-ID, while fully unsupervised person Re-Id has both image-based and video-based applications. Therefore, we summarise and compare unsupervised person Re-ID methods performance in three groups: image-based unsupervised cross-domain person Re-ID, image-based fully unsupervised person Re-ID, and video-based fully unsupervised person Re-ID. We compare unsupervised domain adaptation methods for image-based person Re-ID in Table \ref{table:uda} on four source to target settings, i.e. Duke$\rightarrow$Market, Market$\rightarrow$Duke, Market$\rightarrow$MSMT, and Duke$\rightarrow$MSMT. We summarise image-based fully unsupervised person Re-ID methods performance in Table \ref{table:fully-unsupervised} on four datasets, Market-1501, DueMTMC-Re-ID, MSMT-17 and CUHK03. Unsupervised video-based person Re-ID evaluation results on MARS, DukeMTMC-VideoRe-ID, PRID2011 and iLIDS-VID are summarised in Table \ref{table:video-based-evaluation}.

In image-based unsupervised cross-domain person Re-ID, features alignment methods and GAN-based image transfer methods can't compete with the clustering-based domain adaptation methods. Features alignment methods are more suitable for closed-set application scenarios, where the source and target domains have the same identity labels. However, this is not the case for unsupervised person Re-ID, where the identity labels are unknown in the source and the target domains. The feature alignment approach also assumes that the mid-level features overlap in source and target domains. MMFA and TJ-AIDL rely on mid-level attributes to improve feature transfer from the source to the target domain. However, arbitrary attributes can't capture the dynamics of both source and target domains. For instance, the Market1501 dataset has 27 annotated attributes, while the DukeMTMC-Re-ID dataset only has 23 annotated attributes. Those attributes cannot fully capture many mid-level visual cues. Therefore, UDA models using the mid-level attributes alignment approach can't learn good person feature representations and result in poor evaluation performance. Moreover, attribute annotation incurs labelling costs and shouldn't be considered as unsupervised learning. On the other hand, PAUL~\cite{yang_patch-based_2019} learns mid-level features through body patches and fully exploits the sample space and yields much better performance than MMFA and TJ-AIDL. CASCL~\cite{wu_unsupervised_2019} aligns the mid-level features via intra-camera and inter-camera consistency which further improve the Re-ID performance. D-MMD~\cite{vedaldi_unsupervised_2020-1} and GDS~\cite{vedaldi_global_2020} align features through data distribution. 
pMR-SADA~\cite{wang_smoothing_2020} also uses camera information to alight source feature to target feature and achieve R-1 rate 83\% for Duke$\rightarrow$Market. DAAM~\cite{huang_domain_2020} achieves the best evaluation results in this category by using an attention mechanism only to align source domain-specific features to the target domain while maintaining the target domain-specific features. Therefore, common knowledge is shared between the source and the target while the model is robust on specific target features.

Similarly to the feature alignment approach, image transfer methods such as GAN-based methods preserve source identities in transferred images to supervise model training. The success relies on a great level of similarity in the label space. In really, person identities differ between domains, such as the number of person identities are different across datasets. In addition, the GAN-based image transfer approach only exploits styling difference between the source and target domain without exploiting the full feature dynamics in abundant unlabelled data, therefore, resulting in low performance compare to clustering-based UDA methods, which exploit the full visual similarity in unlabelled data to guide the adaptation procedure. Transferred images are limited to image level discriminative features, which is inferior to local discriminative features. In a multi-camera environment, a person appearance varies significantly due to viewpoint, occlusion etc. therefore, local discriminative features play an essential role in distinguishing a person across views. The GAN-based domain adaptation methods such as PTGAN, SPGAN, CR-GAN, etc. focus on image transfer accuracy from the source to the target domain, ignoring the visual cues in the target domain. The performance of such domain adaptation methods heavily depends on the similar data distribution and the label space. 

Clustering-based methods with pseudo-label refinement coupled with camera variation constraint achieve the best Re-ID accuracy, which explains that camera variations mainly cause domain shift. GLT~\cite{zheng_group-aware_2021} combines the pseudo-label prediction and Re-ID representation learning in one unified optimization objective. The holistic and immediate interaction between these two steps in the training process can significantly help the UDA person Re-ID task.
\begin{table*} [t]
\caption{Image-based fully unsupervised person Re-ID evaluation results.}
\label{table:fully-unsupervised}
\resizebox{\textwidth}{!}{
\begin{tabular}{l|l|rrrr|rrrr|rrrr|rrrr}
\multicolumn{1}{c}{}&\multicolumn{1}{c}{}& \multicolumn{4}{c}{{\bf Market-1501}}& \multicolumn{4}{c}{{\bf
DukeMTMC-Re-ID}}& \multicolumn{4}{c}{{\bf MSMT-17}}& \multicolumn{4}{c}{{\bf CUHK03}}\\
\toprule
{\bf Method} &Year& R-1 & R-5 & R-10 & mAP & R-1 & R-5 & R-10 & mAP & R-1 & R-5 & R-10 & mAP& R-1 & R-5 & R-10 & mAP\\
\midrule
TAUDL~\cite{ferrari_unsupervised_2018}&2018&63.70&-&-&41.20&61.70&-&-&43.50&-&-&-&-&44.70&-&-&31.20\\
BUC~\cite{lin_bottom-up_2019}&2019&66.20&79.60&84.50&38.30&47.40&62.60&68.40&27.50&-&-&-&-&-&-&-&-\\

UGA~\cite{wu_unsupervised_2019-1}&2019&87.20&-&-&70.30&75.00&-&-&53.30&49.50&-&-&21.70&56.50&-&-&68.20\\
DBC~\cite{ding_towards_2019}&2019&69.20&83.00&87.80&41.30&51.50&64.60&70.10&30.00&-&-&-&-&-&-&-&\\
UTAL~\cite{li_unsupervised_2020}&2020&69.20&-&-&46.20&62.30&-&-&44.60&31.40&-&-&13.10&56.30&-&-&42.30\\
TSSL~\cite{wu_tracklet_2020}&2020&71.20&-&-&43.30&45.10&-&-&38.50&-&-&-&-&-&-&-&-\\
SSL~\cite{lin_unsupervised_2020}&2020&71.70&83.80&87.40&37.80&52.50&63.50&68.90&28.60&-&-&-&-&-&-&-&-\\
JVTC~\cite{vedaldi_joint_2020-1}&2020&79.50&89.20&91.90&47.50&74.60&82.90&85.30&50.70&43.10&53.80&59.40&17.30&-&-&-&-\\
HCT~\cite{zeng_hierarchical_2020}&2020&80.00&91.60&95.20&56.40&69.60&83.40&-&87.40&50.70&-&-&-&-&-&-&-\\
MMCL~\cite{wang_unsupervised_2020}&2020&80.30&89.40&92.30&45.50&65.20&75.90&80.00&40.20&35.40&44.80&49.80&11.20&-&-&-&-\\
SpCL~\cite{ge_self-paced_2020}&2020&88.10&95.10&97.00&{\bf 73.10}&-&-&-&-&42.30&55.60&61.20&19.10&-&-&-&-\\
DSCE~\cite{yang_joint_2021}&2021&83.90&92.30&-&61.70&73.80&84.20&-&53.80&35.20&48.30&-&15.50&-&-&-&-\\
IICS~\cite{xuan_intra-inter_2021}&2021&{\bf 89.50}&{\bf 95.20}&{\bf 97.00}&72.90&{\bf 80.00}&{\bf 89.00}&{\bf 91.60}&{\bf 64.40}&{\bf 56.40}&{\bf 68.80}&{\bf 73.40}&{\bf 26.90}&-&-&-&-\\
\bottomrule
\end{tabular}
}
\end{table*}

\begin{table*} [t]
\caption{Video-based fully unsupervised person Re-ID evaluation results.}
\label{table:video-based-evaluation}
\resizebox{\tblwidth}{!}{
\begin{tabular}{l|l|rrrr|rrrr|rrrr|rrrr}
\multicolumn{1}{c}{}&\multicolumn{1}{c}{}&\multicolumn{4}{c}{{\bf MARS}}& \multicolumn{4}{c}{{\bf DukeMTMC-VideoRe-ID}}& \multicolumn{4}{c}{{\bf PRID2011}}& \multicolumn{4}{c}{{\bf iLIDS-VID}}\\
\toprule
{\bf Method} &Year& R-1 & R-5 & R-10 & mAP & R-1 & R-5 & R-10 & mAP& R-1 & R-5 & R-10 & mAP & R-1 & R-5 & R-10 & mAP\\
\midrule
SMP~\cite{liu_stepwise_2017}&2017&23.59&35.81&-&10.54&-&-&-&-&80.90&95.60&98.80&-&41.70&66.30&64.40&-\\
RACE~\cite{ye_robust_2018}&2018&43.20 & 57.10 & 62.10 & 24.50 &-&-&-&-&-&-&-&-&-&-&-&-\\
DAL~\cite{chen_deep_2018}&2018&49.30&65.90&72.20&23.00&-&-&-&-&-&-&-&-&-&-&-&-\\
TAUDL~\cite{ferrari_unsupervised_2018}&2018&43.80&59.90&-&29.10&-&-&-&-&49.40&78.70&-&-&26.70&51.30&-&-\\
BUC~\cite{lin_bottom-up_2019}&2019&61.10&75.10&80.00&38.00&69.20&81.10&85.80&61.90&-&-&-&-&-&-&-&-\\
UGA~\cite{wu_unsupervised_2019-1}&2019&58.10&73.40&-&39.30&-&-&-&-&80.90&94.40&-&-&57.30&72.00&-&-\\
UTAL~\cite{li_unsupervised_2020}&2020&49.90&66.40&-&35.20&-&-&-&-&54.70&83.10&-&-&35.10&59.00&-&-\\
TSSL~\cite{wu_tracklet_2020}&2020&56.30&-&-&30.50&73.90&-&-&64.60&-&-&-&-&-&-&-&-\\
SSL~\cite{lin_unsupervised_2020}&2020&{\bf 62.80}&{\bf 77.20}&{\bf 80.10}&{\bf 43.60}&{\bf 76.40}&{\bf 88.70}&{\bf 91.00}&{\bf 69.30}&-&-&-&-&-&-&-&-\\
\bottomrule
\end{tabular}
}
\end{table*}
We compare image-based fully unsupervised person Re-ID methods performance on the Market-1501, DukeMTMC-Re-ID, MSMT-17 and CUHK03 datasets in Table \ref{table:fully-unsupervised}. Fully unsupervised person Re-ID is more challenging than unsupervised cross-domain person Re-ID as there is no labelled source data. The most common approach is to cluster unlabelled data into many clusters based on visual similarity to obtain pseudo-labels to supervise model tanning. Clustering on visual similarity is not sufficient to learn cross-camera invariant features because intra-camera people always share greater similarity due to common background, lighting conditions, etc. To enhance cross-camera invariance, several methods exploit camera information to achieve better results. In particular, IICS leverages the intra-camera and inter-camera similarities in learning cross-camera invariance, which achieves the state-of-the-art evaluation results on Market-1501, DukeMTMC-Re-ID and MSMT17 datasets.  

In Table \ref{table:uda} and Table \ref{table:fully-unsupervised}, performance on the MSMT17 dataset is significantly worse than on the Market1501 and DukeMTMC-Re-ID datasets in both unsupervised cross-domain and fully unsupervised person Re-ID settings. The MSMT17 dataset is more challenging than the Market1501 and DukeMTMC-Re-ID datasets because of more complex lighting, scene variations, random distracting samples. In addition, MSMT17 has a much larger number of identities (4,101) than Market (1,501) and Duke (1,404). Unsupervised cross-domain person Re-Id models learnt on source domains are sensitive to person feature variations in unknown target domains, which explains why UDA methods achieve better evaluation results on the Duke and Market datasets. Similarly, image-based fully unsupervised person Re-ID models perform better on the Market and the Duke datasets than on the MSMT17 datasets because it is more challenging to learn person feature representations for 4,101 identities with more variations than for 1,501 and 1,404 identities with fewer variations.

Video-based fully unsupervised person Re-ID methods performance is summarised in Table \ref{table:video-based-evaluation}. Video-based unsupervised person Re-ID is a less investigated area. Most video-based unsupervised learning uses temporary information in tracklets to associate person identities to tracklets. The clustering approach is commonly used to generate pseudo-labels for the tracklets. To address the impact of cross-camera hard samples, SSL~\cite{lin_unsupervised_2020} learns soft labels for training samples, which is particularly helpful with the hard samples, which yields the top performance.

\section{Discussion and Future Directions}\label{future}
This survey reviews the recent Re-ID advances focusing on unsupervised methods that address the data scalability issue. Despite the fact that remarkable achievements have been made in the past few years, it remains an open question on addressing the significant person Re-ID challenges, namely the identity label estimation, discriminative person feature learning, camera-invariant person feature learning and the domain gap between datasets. Next, we highlight a few promising future research directions.

\paragraph{Unsupervised end-to-end person Re-ID.}While most person Re-ID works use hand-annotated or detected bounding boxes to train the Re-ID, it is more practical to learn a person Re-Id model end-to-end in which raw images and video are used directly to train the Re-ID model. ~\cite{xu_person_2014} proposed to combine the person detection and person Re-ID in a unified process. Since then, several supervised end-to-end person re-identification methods have been proposed and proved that jointly considering person detection and person Re-ID leads to higher retrieval accuracy than using them separately due to the detection-identification inconsistency issue~\cite{ijcai2021-613}. However, there is still a gap in utilizing large unlabelled images and videos to learn Re-ID models in an unsupervised end-to-end manner, which is more practical in real-world applications. Therefore, we suggest there is a great research opportunity in unsupervised end-to-end person re-identification, in particular, leveraging the evolutionary vision transformers~\cite{khan_transformers_2021,10.1007/978-3-030-58452-8_13,nguyen_clusformer_2021,liu_swin_2021}.

\paragraph{Unsupervised generalizable person Re-ID.}
Unsupervised cross-domain person Re-Id models assume that the source and target domains share considerable data similarity. Thus the knowledge learnt in the source domain can be adapted into the target domain. However, this adaptation process is excessive and is required on each target domain, which restricts their applicability in arbitrarily domains~\cite{ferrari_unsupervised_2018}. The success of unsupervised cross-domain person Re-Id models depends on the source domain's prior knowledge of the target domain. Therefore, when the source domain data are diverse enough that cover the target domain, the pre-trained Re-ID model on the source domain may be able to deploy to the target domain without domain adaptation. Namely, unsupervised generalizable person Re-ID is achieved. A generalizable Re-Id model must be robust and discriminative in multiple environments. Therefore the model needs to be trained with sufficient annotated and synthesized data covering diverse photometric and geometric person variations.  DIMN~\cite{song_generalizable_2019} and OsNet~\cite{zhou_omni-scale_2019} attempted to learn one-fit-all person feature representations out of public labelled data in a supervised manner. Person representations are averaged over multiple representations hoping that they would work in the unseen domains. However, these models didn't perform well as they didn't see the data distributions in the target domains explored. Labelling cost is the biggest hurdle to collect sufficient annotated data to train models that generalize well to unknown domains. Instead, unlabelled data are easy to collect and will cover more variations for the Re-ID purpose. Therefore, unsupervised generalizable person re-identification is a promising direction due to the benefit of zero deployment cost to arbitrary domains.

\paragraph{Unsupervised text-based cross-domain person Re-ID.}
Text-based person re-identification ~\cite{li_person_2017b} aims to find a target person from a gallery of images or videos for a given textual description. Text-based Re-ID is handy when the query image is absent while the textual description of the target person is available. It's precious in an emergency environment such as the missing person search. While training a text-based Re-ID model for every environment is prohibitive, it's reasonable to pre-train a model with annotated data and adapt to unlabeled domains. Like the image-based unsupervised cross-domain person Re-ID, the abundant unlabelled data can be exploited to learn person feature representations for the target domains. Instead of predicting global pseudo-labels, the adaptation process needs to align word-level features between the source and the target domains for text-based domain adaptation. The attention mechanism and various distribution distance losses can be employed to facilitate feature representation learning.

\section{Conclusion}\label{conclusion}

This survey delivers a comprehensive review of the recent works on unsupervised person re-identification, which address the data scalability problem. Unsupervised person Re-ID has attracted much attention due to its practical applications and research significance. For the first time, we surveyed unsupervised person Re-ID methods in both image-based and video-based applications. We discuss highly regarded methods from the perspective of challenges and solutions to inspire new ideas. We summarise and analyze unsupervised person Re-ID methods performance. We provide insights that unsupervised person Re-ID methods need to address the joint challenge of identity label estimation, discriminative person feature representation learning, cross-camera invariant feature learning and domain gap between source and target datasets. We finally suggest several promising future research directions to inspire new research in the field.

\printcredits

\section*{Declaration of competing interest}
The authors declare that they have no known competing financial interests or personal relationships that could have appeared to influence the work reported in this paper.

\section*{Acknowledgement}
This work is partially supported by Australian Research Council (ARC) Discovery Early Career Researcher Award (DECRA) under DE190100626.

\bibliographystyle{elsarticle-num}

\bibliography{bib}



\end{document}